\documentclass[twocolumn,twoside]{IEEEtran}    % TWOCOLS

\usepackage[ansinew]{inputenc}
\usepackage{color}
\usepackage{array}
\usepackage{colortbl}
\usepackage{amsmath}
\usepackage{amssymb}
\usepackage{float}
\usepackage{amscd}
\usepackage{amsfonts}
\usepackage{amssymb}
\usepackage{cite}
\usepackage{psfrag}
\usepackage[dvips]{graphicx}
\definecolor{Gray}{gray}{0.95}

\usepackage[caption=false, font=footnotesize]{subfig}

\newcommand{\x}{{\mathbf x}}

\newcommand{\y}{{\mathbf y}}
\newcommand{\Y}{{\mathbf Y}}
\newcommand{\X}{{\mathbf X}}

\newcommand{\W}{{\mathbf W}}

\newcommand{\I}{{\mathbf I}}
\newcommand{\U}{{\mathbf U}}
\newcommand{\V}{{\mathbf V}}
\newcommand{\C}{{\mathbf C}}

\newcommand{\uu}{{\mathbf u}}
\newcommand{\vv}{{\mathbf v}}

\newcommand{\A}{{\mathbf A}}

\newcommand{\PHI}{\boldsymbol{\Phi}}
\newcommand{\PHi}{\boldsymbol{\phi}}

\newcommand{\aalpha}{\boldsymbol{\alpha}}
\newcommand{\BETA}{\boldsymbol{\beta}}

\def\argmin{\mathop{\rm arg~min}}

\newcommand{\IG}{\includegraphics}
\newcommand{\Real}{\mathbb R}

\newcommand{\Kbf}{\mathbf{K}}

\newcommand{\Rbf}{\mathbf{R}}

\title{Kernel Multivariate Analysis Framework for Supervised Subspace Learning: A Tutorial on Linear and Kernel Multivariate Methods}
\author{Jer\'onimo Arenas-Garc\'ia,~\IEEEmembership{Senior Member,~IEEE}, Kaare Brandt Petersen, \\Gustavo Camps-Valls,~\IEEEmembership{Senior Member,~IEEE}, and Lars Kai Hansen
}

\begin{document}

\maketitle

\vspace{-1.5cm}
{\small
\noindent {\bf Abstract:} Feature extraction and dimensionality reduction are important tasks in many fields of science dealing with signal processing and analysis. The relevance of these techniques is increasing as current sensory devices are developed with ever higher resolution, and problems involving multimodal data sources become more common. A plethora of feature extraction methods are available in the literature collectively grouped under the field of Multivariate Analysis (MVA). This paper provides a uniform treatment of several methods: Principal Component Analysis (PCA), Partial Least Squares (PLS), Canonical Correlation Analysis (CCA) and Orthonormalized PLS (OPLS), as well as their non-linear extensions derived by means of the theory of reproducing kernel Hilbert spaces. We also review their connections to other methods for classification and statistical dependence estimation, and introduce some recent developments to deal with the extreme cases of large-scale and low-sized problems. To illustrate the wide applicability of these methods in both classification and regression problems, we analyze their performance in a benchmark of publicly available data sets, and pay special attention to specific real applications involving audio processing for music genre prediction and hyperspectral satellite images for Earth and climate monitoring. }
%\end{abstract}}

\IEEEpeerreviewmaketitle

\section{Introduction}

As sensory devices develop with ever higher resolution and the combination of diverse data sources is more common, feature extraction and dimensionality reduction become increasingly important in automatic learning. This is especially true in fields dealing with intrinsically high-dimensional signals, such as those acquired for image analysis, spectroscopy, neuroimaging, and remote sensing, but also for situations when many heterogeneous features are computed from a signal and stacked together for classification, clustering, or prediction.

Multivariate analysis (MVA) constitutes a family of methods for dimensionality reduction successfully used in several scientific areas~\cite{Wold66b}. The goal of MVA algorithms is to exploit correlations among the variables to find a reduced set of features that are relevant for the learning task. Among the most well-known MVA methods are Principal Component Analysis (PCA), Partial Least Squares (PLS), and Canonical Correlation Analysis (CCA). PCA disregards the target data and exploits correlations between the input variables to maximize the variance of the projections, while PLS and CCA look for projections that maximize, respectively, the covariance and the correlation between the features and the target data.  Therefore,  they should in principle be preferred to PCA for regression or classification problems. In this paper, we consider also a fourth MVA method known as Orthonormalized PLS (OPLS) that is also well-suited to supervised problems, with certain optimality in least-squares (LS) multiregression. A common advantage of these methods is that they can be formulated using standard linear algebra, and can be implemented as standard or generalized eigenvalue problems. Furthermore, implementations and variants of these methods have been proposed that operate either in a blockwise or iterative manner to improve speed or numerical stability.

No matter how refined the various methods of MVA are, they are still constrained to account for linear input-output relations. Hence, they can be severely challenged when features exhibit non-linear relations. In order to address these problems, non-linear versions of MVA methods have been developed and these can be classified into two fundamentally different approaches~\cite{Rosipal11}: 1) The modified methods in which the linear relations among the latent variables are substituted by non-linear relations~\cite{Woldetal89,Qin92}; and 2) Variants in which the algorithms are reformulated to fit a kernel-based approach~\cite{Boser92,scholkopf1998,ShaweTaylor04}. In this paper, we will review the second approach, in which the input data is mapped by a non-linear function into a high-dimensional space where the ordinary MVA problems are stated. A central property of this kernel approach is the exploitation of the so-called {\em kernel trick}, by which the inner products in the transformed space are replaced with a kernel function working solely with input space data so the explicit non-linear mapping is not explicitly necessary. 
\begin{table*}[t!]
\begin{center}
\caption{Acronyms and notation used in the paper.\label{tab_acro}}
\renewcommand{\baselinestretch}{1.2} \footnotesize
\setlength{\tabcolsep}{4pt}
\begin{tabular}{ll|ll}
\hline
\hline
AR       &   Autoregressive & $l$ & Size of labelled dataset \\
ECMWF    &   European Centre for Medium-Range Weather Forecasts& $u$ & Number of unlabeled data\\ 
GMM      &   Gaussian Mixture Model& $n = l + u$ & Total number of training samples\\ 
HSIC     &   Hilbert-Schmidt Independence Criterion & $d$& Dimension of input space\\ 
IASI     &   Infrared Atmospheric Sounding Interferometer& $m$ & Dimension of output space \\ 
(k)CCA     &   (Kernel) Canonical Correlation Analysis  & $\X$& Centered input data (size $l \times d$) \\ 
kFD      &   Kernel Fisher Discriminant& $\Y$& Centered output data (size $l \times m$) \\ 
(k)MVA     &   (Kernel) Multivariate Analysis& $\C_x, \C_y $ & Input, output data sample covariance matrices\\ 
(k)OPLS    &   (Kernel) Orthonormalized Partial Least Squares& $\C_{xy} $ & Input-output sample cross-covariance matrix \\ 
(k)PCA     &   (Kernel) Principal Component Analysis& $\X^\dag$ & Moore-Penrose pseudoinverse of matrix $\X$ \\ 
(k)PLS     &   (Kernel) Partial Least Squares& $\W$ & Regression coefficients matrix \\ 
MSE      &   Mean-square error&  $\I$ & Identity matrix \\ 
LDA      &   Linear Discriminant Analysis& $\|\mathbf{A}\|_F$ & Frobenius norm of matrix $\mathbf A$\\ 
LS       &   Least-squares& $n_f$ & Number of extracted features \\ 
OA       &   Overall Accuracy& $\uu_i $, $\vv_i$ & $i$th projection vector for the input, output data \\ 
RBF      &   Radial Basis Function& $\U$, $\V$ & $[\uu_1,\dots,\uu_{n_f}]$, $[\vv_1,\dots,\vv_{n_f}]$. Projection matrices \\ 
rkCCA    &   reduced complexity kCCA& $\X'$, $\Y'$ & Extracted features for the input, output data \\ 
rkOPLS   &   reduced complexity kOPLS & $\cal F$ & Reproducing kernel Hilbert Space \\
rkPCA    &   reduced complexity kPCA& $\PHi(\x)$ & Mapping of $\x$ in feature space \\ 
RMSE     &   Root-mean-square Error& $k(\x_i,\x_j)$ & $\langle \PHi(\x_i), \PHi(\x_j)\rangle_{\cal F}$. Kernel function \\ 
RTM      &   Radiative transfer models& $\PHI$ & $[\PHi(\x_1),\dots,\PHi(\x_l)]^\top$. Input data in feature space \\ 
skPLS    &   Sparse kPLS&   $\mathbf{K}_x = \PHI \PHI^\top$ & Gram Matrix \\ 
UCI      &   University of California, Irvine& ${\mathbf{A}}$ & $[\aalpha_1, \cdots, \aalpha_{n_f}]$. Coefficients for $\U = \PHI^\top \mathbf{A} $ \\ 
\hline
\hline
\vspace{-1cm}
\end{tabular}
\end{center}
\end{table*}

An appealing property of the resulting kernel algorithms is that they obtain the flexibility of non-linear expressions using straightforward methods from linear algebra. However, supervised kernel MVA methods are hampered in applications involving  large datasets or a small number of labeled samples. Sparse and incremental versions have been presented to deal with the former problem, while the field of semisupervised learning has recently emerged for the latter. Remedies to these problems involve a particular kind of {\em regularization}, guided either by selection of a reduced number of basis functions or by considering the information about the manifold conveyed by the unlabeled samples. Both approaches will be also reviewed in this paper. Concretely, we aim:
\begin{enumerate}
\item To review linear and kernel MVA algorithms, providing their theoretical characterization and comparing their main properties under a common framework.
\item To present relations between kernel MVA and other feature extraction methods based on Fisher's discriminant analysis and nonparametric kernel dependence estimates.
\item To review sparse and semisupervised approaches that make the kernel variants practical for large-scale or undersampled labeled datasets. We will illustrate how these approaches overcome some of the difficulties that may have limited a more widespread use of the most powerful kernel MVA methods.
\item To illustrate the wide applicability of these methods, for which we consider several publicly available data sets, and two real scenarios involving audio processing for music genre prediction and hyperspectral satellite images for Earth and climate monitoring. Methods will be assessed in terms of accuracy and robustness to the number of extracted features.
\end{enumerate}

We continue the paper with a brief review of linear and kernel MVA algorithms, as well as connections to other methods. Then, Section~\ref{sec:sparsesemi} introduces some extensions that increase the applicability of kernel MVA methods in real applications. Section~\ref{sec:experiments} provides illustrative evidence of method's performance. Finally, we conclude the paper in Section~\ref{sec:conclusions} with some discussion and future lines of research.

\section{Multivariate Analysis in Reproducing Kernel Hilbert Spaces}\label{sec:mvarkhs}

This section reviews the framework of MVA both in the linear case and with kernel methods. Interesting connections are also pointed out between particular classification and dependence estimation kernel methods. Table~\ref{tab_acro} provides a list of acronyms, and basic notation and variables that will be used throughout the paper.

\subsection{Problem statement and notation}

Let us consider a supervised regression or classification problem, and let $\X$ and $\Y$ be columnwise-centered input and target data matrices of sizes $l\times d$ and $l \times m$, respectively. Here, $l$ is the number of training data points in the problem, and $d$ and $m$ are the dimensions of the input and output spaces, respectively. The target data can be either a set of variables that need to be approximated, or a matrix that encodes the class membership information. The sample covariance matrices are given by $\mathbf{C}_{x} = \frac{1}{l} \X^\top \X$ and $\mathbf{C}_{y} = \frac{1}{l} \Y^\top \Y$, whereas the covariance between the input and output data is $\mathbf{C}_{xy} = \frac{1}{l} \X^\top \Y$.

The objective of standard linear multiregression is to adjust a linear model for predicting the output variables from the input features, $\widehat{\Y} = \X \mathbf{W}$, where $\mathbf{W}$ contains the regression model coefficients. Ordinary least-squares (LS) regression solution is $\mathbf{W} = \X^\dagger \Y$, where $\X^\dagger = (\X^\top \X)^{-1} \X^\top$ is the Moore-Penrose pseudoinverse of $\X$. Highly correlated input variables can result in rank-deficient $\mathbf{C}_x$, making the inversion of this matrix unfeasible. The same situation is encountered in the small-sample-size case (i.e., when $l<d$, which is usually the case when using kernel extensions). Including a Tikhonov's regularization term leads to better conditioned solutions: for instance, by also minimizing the Frobenius norm of the weights matrix $\|\mathbf{W}\|_F^2$, one obtains the regularized LS solution $\mathbf{W} = (\X^\top \X + \lambda{\bf I})^{-1} \X^\top\Y$, where parameter $\lambda$ controls the amount of regularization.

The solution suggested by MVA to the above problem consists in projecting the input data into a subspace that preserves the most relevant information for the learning problem. Therefore, MVA methods obtain a transformed set of features via a linear transformation of the original data, $\X' = \X\U$, where $\U = [\uu_1, \uu_2, \dots, \uu_{n_f}]$ will be referred hereafter as the projection matrix, $\mathbf{u}_i$ being the $i$th projection vector and $n_f$ the number of extracted features\footnote{Note that strictly speaking $\U$ is not a projection operator, since it implies a transformation from $\Real^d$ to $\Real^{n_f}$ and does not satisfy the idempotent property of projection operators. Nevertheless, if the columns of $\U$ are linearly independent, vectors $\uu_i$ constitute a basis of the subspace of $\Real^d$ where the data is projected, and it is in this sense that we refer to $\uu_i$ and $\U$ as projection vectors and matrix, and to ${\X'=\X\U}$ as projected data. This nomenclature has been widely adopted in the machine learning field~\cite{SPM11}.}. Some MVA methods consider also a feature extraction in the output space, $\Y' = \Y \V$, with $\V = [\vv_1, \vv_2, \dots, \vv_{n_f}]$.

Generally speaking, MVA methods look for projections of the input data that are ``maximally aligned'' with the targets, and the different methods are characterized by the particular objectives they maximize. Table~\ref{properties} provides a summary of the MVA methods that we will discuss in the rest of the section. An interesting property of linear MVA methods is that they are based on first and second order moments, and that their solutions can be formulated in terms of (generalized) eigenvalue problems. Thus, standard linear algebra methods can be readily applied.

\begin{table*}[t!]
\small
\caption{\label{properties}Summary of linear and kernel MVA methods. For each method it is stated the objective to maximize (1st row), constraints for the optimization (2nd row), and maximum number of features (last row).}
\begin{center}
\setlength{\tabcolsep}{4pt}
\begin{tabular}{cccccccc}
\hline
\hline
{\bf PCA} & {\bf PLS} & {\bf CCA} & {\bf OPLS} & {\bf kPCA} & {\bf kPLS} & {\bf kCCA} & {\bf kOPLS} \\
\hline
\hline
\scriptsize$ \uu^\top \C_{x} \uu$ &  {\scriptsize$\uu^\top \C_{xy} \vv$} & \scriptsize$\uu^\top \C_{xy}  \vv$ & \scriptsize$\uu^\top \C_{xy} \C_{xy}^\top \uu$ & \scriptsize$ \aalpha^\top \Kbf_x^2 \aalpha$ & {\scriptsize$\aalpha^\top \Kbf_x \Y \vv$} & \scriptsize$\aalpha^\top \Kbf_x \Y  \vv$ & \scriptsize$\aalpha^\top \Kbf_x \Y\Y^{\top} \Kbf_x \aalpha$\\
\hline
\scriptsize$\U^\top \U = \I$ & {\scriptsize$\begin{array}{c} \U^\top \U = \I \\ \V^\top \V = \I  \end{array}$} & \scriptsize$\begin{array}{c} \U^\top \C_{x} \U = \I \\ \V^\top \C_{y} \V = \I \end{array}$ & \scriptsize$\U^\top \C_{x} \U = \I$ & \scriptsize$\A^\top \Kbf_x \A = \I$ & {\scriptsize$\begin{array}{c} \A^\top \Kbf_x \A = \I \\ \V^\top \V = \I  \end{array}$} & \scriptsize$\begin{array}{c} \A^\top \Kbf_x^2 \A = \I \\ \V^\top \C_{y} \V = \I \end{array}$ & \scriptsize$\A^\top \Kbf_x^2 \A = \I$\\
\hline
\scriptsize $r(\X)$ & \scriptsize $r(\X)$ & \scriptsize $r(\C_{xy})$ & \scriptsize $r(\C_{xy})$ & \scriptsize $r(\Kbf_x)$ & \scriptsize $r(\Kbf_x)$ & \scriptsize $r(\Kbf_x\Y)$ & \scriptsize $r(\Kbf_x\Y)$ \\
\hline
\hline
\end{tabular}
\end{center}
\footnotesize Vectors ${\bf u}$ and $\aalpha$ are column vectors in matrices ${\U}$ and ${\bf A}$, respectively. $r(\cdot)$ denotes the rank of a matrix.
\vspace{-.5cm}
\end{table*}

\subsection{Linear Multivariate Analysis}

Principal Component Analysis, which is also known as the \emph{Hotelling transform} or the \emph{Karhunen-Lo\`eve transform}~\cite{Jollife86}, is probably the most widely used MVA method and the oldest dating back to 1901~\cite{pearson1902}. PCA selects the maximum variance projections of the input data, imposing an orthonormality constraint for the projection vectors (see Table~\ref{properties}). PCA works under the hypothesis that high variance projections contain the relevant information for the learning task at hand. PCA is based on the input data alone, and is therefore an {\em unsupervised} feature extraction method. Methods that explicitly look for the projections that better explain the target data should in principle be preferred in a supervised setup. Nevertheless, PCA and its kernel version, kPCA, are used as a preprocessing stage in many supervised problems, likely because of their simplicity and ability to discard irrelevant directions~\cite{Braun08,Abrahansem11}.

The PLS algorithm~\cite{wold1966a} is based on  latent variables that account for the information in $\mathbf{C}_{xy}$. In order to do so, PLS extracts the projections that maximize the covariance between the projected input and output data, again under orthonormality constraints for the projection vectors. This is done either as an iterative procedure or by solving an eigenvalue problem. In the iterative schemes, the data sets $\mathbf{X}$ and $\mathbf{Y}$ are recursively transformed in a process which subtracts the information contained in the already estimated latent variables. This process, which is often referred to as \emph{deflation}, can be done in a number of ways that define the many variants of PLS. Perhaps the most popular PLS method was presented in~\cite{wold1984}. The algorithm, hereafter referred to as PLS2, assumes a linear relation between $\mathbf{X}$ and $\mathbf{Y}$ that implies a certain deflation scheme, where the latent variable of $\mathbf{X}$ is used to deflate also $\mathbf{Y}$ \cite[p. 182]{ShaweTaylor04}. Several other variants of PLS exist such as `PLS Mode A' and PLS-SB; see~\cite{geladi1988} for a discussion of the early history of PLS and~\cite{kramer2006} for a well-written contemporary overview.
Among its many advantages, PLS does not involve matrix inversion and deals efficiently  with highly correlated data. This has justified its very extensive use in fields such as chemometrics and remote sensing, where signals typically are acquired in a range of highly correlated spectral wavelengths.

Rather than maximizing covariance, CCA maximizes the correlation between projected input and output data~\cite{Hotelling36}. In this way, CCA can more conveniently deal with directions of the input or output spaces that present very high variance, and that would therefore be over-emphasized by PLS, even if the correlation between the projected input and output data is not very significant.

A final method we will pay attention to is Orthonormalized PLS (OPLS), also known as multilinear regression~\cite{Borga97} or semi-penalized CCA~\cite{Barker03}. OPLS is optimal for performing multilinear LS regression on the features extracted from the training data, i.e., 
\begin{equation}\U = \displaystyle\mathop{\argmin}_{\U} \|\Y - \X' \mathbf{W}\|_F^2.
\end{equation}
with $\W = \X'^\dag \Y$ being the matrix containing the optimal regression coefficients. It can be shown that this optimization problem is equivalent to the one stated in Table~\ref{properties}. Alternatively, this problem can also be associated  with the maximization of a Rayleigh coefficient involving projections of both input and output data, $\frac{(\uu^\top {\mathbf C}_{xy} \vv)^2}{(\uu^\top {\mathbf C}_{x} \uu)(\vv^\top \vv)}$.  It is in this sense that this method is called semi-penalized CCA, since it disregards variance of the projected input data, but rewards those input features that better predict large variance projections of the target data. This asymmetry makes sense in supervised subspace learning where matrix $\mathbf{Y}$ contains target values to be approximated from the extracted input features. In fact, it has been shown that for classification problems OPLS is equivalent to Linear Discriminant Analysis (LDA), provided an appropriate labeling scheme is used for $\Y$~\cite{Barker03}. However, in ``two-view learning'' problems in which $\mathbf{X}$ and $\mathbf{Y}$ represent different views of the data \cite[Sec. 6.5]{ShaweTaylor04}, one would like to extract features that can predict both data representations simultaneously, and CCA could be preferred to OPLS.

A common framework for PCA, PLS, CCA and OPLS was proposed in~\cite{Borga97}, where it was shown that these methods can be reformulated as (generalized) eigenvalue problems, so that linear algebra packages can be used to solve them. Concretely:
\begin{eqnarray}
%\small
\arraycolsep=2pt
\begin{array}{ll}
{\rm PCA:} &\mathbf{C}_{x} {\uu} = \lambda \uu \\[5mm]
 {\rm PLS:} & \left(\begin{array}{ll}\mathbf{0} & \mathbf{C}_{xy} \\ \mathbf{C}_{xy}^\top & \mathbf{0}\end{array}\right) \left(\begin{array}{c}\uu \\ \vv \end{array}\right) = \lambda \left(\begin{array}{c}\uu \\ \vv \end{array}\right) \\ [5mm]
{\rm OPLS:} & \mathbf{C}_{xy}\mathbf{C}_{xy}^\top {\uu} = \lambda \mathbf{C}_{x}\uu \quad \\[5mm]
  {\rm CCA:} & \left(\begin{array}{ll}\mathbf{0} & \mathbf{C}_{xy} \\ \mathbf{C}_{xy}^\top & \mathbf{0}\end{array}\right) \left(\begin{array}{c}\uu \\ \vv \end{array}\right) = \lambda  \left(\begin{array}{ll}\mathbf{C}_{x} & \mathbf{0}  \\ \mathbf{0} & \mathbf{C}_{y} \end{array}\right) \left(\begin{array}{c}\uu \\ \vv \end{array}\right)
\end{array}
\end{eqnarray}

We can see that CCA and OPLS require the inversion of matrices $\mathbf{C}_{x}$ and $\mathbf{C}_{y}$. If these are rank-deficient, then it becomes necessary to first extract the dimensions with non-zero variance using PCA, and then solve the CCA or OPLS problems. A very common approach is to solve the above problems using a two-step iterative procedure. In the first step, the projection vectors corresponding to the largest (generalized) eigenvalue are chosen, for which there exist efficient methods such as the power method. The second step is known as {\em deflation}, and consists in removing from the data or covariance matrices the variance that can be already explained with the features extracted in the first step. Alternative solutions for these methods can be obtained by reformulating them as regularized least squares minimization problems. For instance, the work in~\cite{Zou06,Hardoon11,Heskes12} introduced sparse versions of PCA, CCA and OPLS by adding sparsity promotion terms, such as LASSO or $\ell_1$-norm on the projection vectors, to the LS functional.

\begin{figure*}[t!]
\centerline{\includegraphics[width=18cm,trim=2cm 10cm 0 10cm]{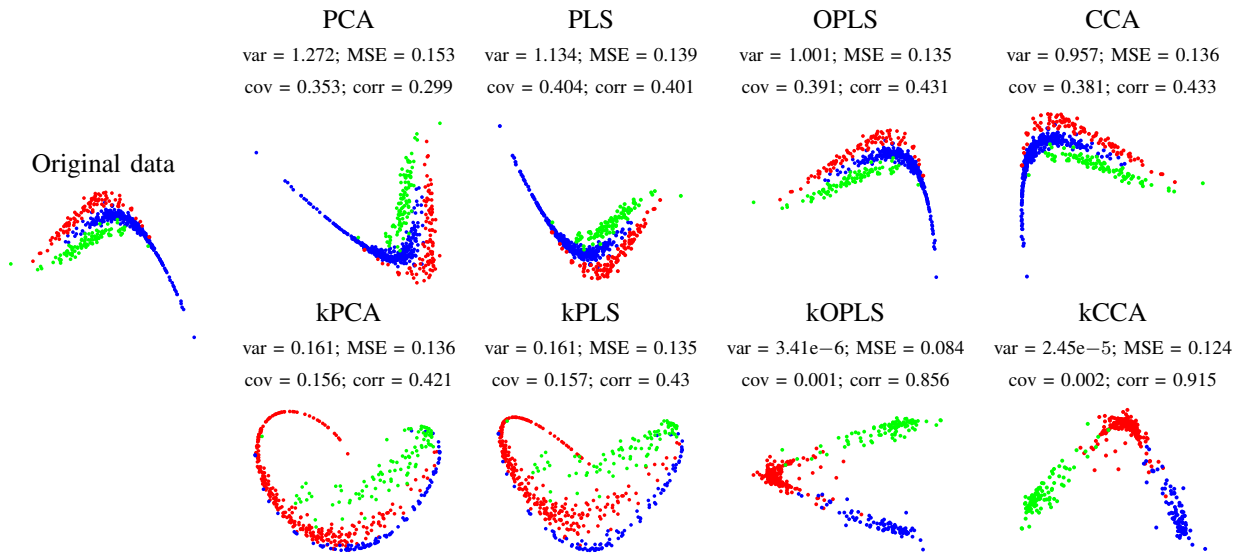}}
\caption{Features extracted by different MVA methods in a three-class problem. For the first feature extracted by each method we show its variance (var), the mean-square-error when the projected data is used to approximate $\mathbf{Y}$ (MSE), and the largest covariance (cov) and correlation (corr) that can be achieved with any linear projection of the target data. \label{fig1}}
\vspace{-.5cm}
\end{figure*}

Figure~\ref{fig1} illustrates the features extracted by the methods for a toy classification problem with three classes. The data was generated from three noisy sinusoid fragments, so that a certain overlap exists between classes. For the application of supervised methods, class membership is defined in matrix $\Y$ using 1-of-$c$ encoding~\cite{Bishop95}. Above each scatter plot we provide, for the first extracted feature, its sample variance, the largest covariance and correlation that can be achieved with any linear transformation of the output data, and the optimum mean-square-error (MSE) when that feature is used to approximate the target data, i.e., $\frac{1}{l m}\|\Y - \X \uu_1 (\X\uu_1)^\dag \Y \|_F^2$. As expected, the first projection of PCA, PLS and CCA maximize, respectively, the variance, covariance and correlation, whereas OPLS finds the projection that minimizes the MSE. However, since these methods can just perform linear transformations of the data, they are not able to capture any non-linear relations between the input variables.

\subsection{Kernel Multivariate Analysis}
The framework of kernel MVA (kMVA) algorithms is aimed at extracting nonlinear projections while actually working with linear algebra. Let us first consider a function $\PHi: \Real^d \rightarrow \cal F$ that maps input data into a Hilbert feature space $\cal F$. The new mapped data set is defined as $\PHI = [\PHi(\x_1), \cdots, \PHi(\x_l)]^\top$, and the features extracted from the input data will now be given by $\PHI' = \PHI \U$, where matrix $\U$ is of size $\text{dim}({\cal F}) \times n_f$. The direct application of this idea suffers from serious practical limitations when the dimension of $\cal F$ is very large, which is typically the case.

To implement practical kernel MVA algorithms we need to rewrite the equations in the first half of Table~\ref{properties} in terms of inner products in $\cal F$ only. For doing so, we rely on the availability of a kernel matrix $\Kbf_x = \PHI  \PHI^\top$ of dimension $l \times l$, and on the Representer's Theorem~\cite{ShaweTaylor04}, which states that the projection vectors can be written as a linear combination of the training samples, i.e, $\U = \PHI^\top  \A$, matrix $\A = [\aalpha_1, \dots,\aalpha_{n_f}]$ being the new argument for the optimization\footnote{In this paper, we assume that data is centered in feature space, what can easily be done through a simple modification of the kernel matrix \cite{ShaweTaylor04}.} This is typically referred to as the {\em kernel trick} and has been used to develop kernel versions of the previous linear MVA, as indicated in the last four columns of Table~\ref{properties}.
 
For PCA, it was Sch\"olkopf, Smola and M\"uller who in 1998 introduced a kernel version denoted kPCA~\cite{scholkopf1998}. Lai and Fyfe in 2000 first introduced the kernel version of CCA denoted kCCA~\cite{lai2000} (see also~\cite{ShaweTaylor04}). Later, Rosipal and Trejo presented a non-linear kernel variant of PLS in~\cite{Rosipal01}. In that paper, $\Kbf_x$ and the $\mathbf{Y}$ matrix are deflated the same way, which goes more in line with the PLS2 variant than to the traditional algorithm `PLS Mode A', and therefore we will denote it as kPLS2. A kernel variant of Orthonormalized PLS was presented in~\cite{ArenasGarcia07} and is here referred to as kOPLS. Specific versions of kernel methods to deal with signal processing applications have also been proposed, such as the temporal kCCA of~\cite{Biessmann10}, that is designed to exploit temporal structure in the data.

As for the linear case, kernel MVA methods can be implemented as (generalized) eigenvalue problems:
\begin{eqnarray}
%\small
\arraycolsep=2pt
\begin{array}{ll} \label{kernelitos}
{\rm kPCA:} & \mathbf{K}_{x} {\aalpha} = \lambda \aalpha \\[5mm] 
{\rm kPLS:} & \left(\begin{array}{ll}\mathbf{0} & \mathbf{K}_{x} \Y \\ \Y\mathbf{K}_{x} & \mathbf{0}\end{array}\right) \left(\begin{array}{c}\aalpha \\ \vv \end{array}\right) = \lambda \left(\begin{array}{c}\aalpha \\ \vv \end{array}\right) \\[5mm] 
{\rm kOPLS:} &\mathbf{K}_{x}\Y\Y^\top\mathbf{K}_{x} {\aalpha} = \lambda \mathbf{K}_{x} \mathbf{K}_{x}\aalpha \quad \\[5mm]
 {\rm kCCA:~} & \left(\begin{array}{ll}\mathbf{0} & \mathbf{K}_{x} \Y \\ \Y\mathbf{K}_{x} & \mathbf{0}\end{array}\right) \left(\begin{array}{c}\aalpha \\ \vv \end{array}\right) = \lambda  \left(\begin{array}{ll}\mathbf{K}_{x}\mathbf{K}_{x} & \mathbf{0}  \\ \mathbf{0} & \mathbf{C}_{y} \end{array}\right) \left(\begin{array}{c}\aalpha \\ \vv \end{array}\right)  
\end{array}
\end{eqnarray}
It should be noted that the output data could also be mapped to some feature space $\cal H$, as it was considered for kCCA in~\cite{lai2000} for a multi-view learning case. Here, we consider that it is the actual labels in $\Y$ which need to be well-represented by the extracted input features, so we will deal with the original representation of the output data.

For illustrative purposes, we have incorporated to Fig. 1 the projections obtained in the toy problem by kMVA methods using the {\em radial basis function} (RBF) kernel, $k(\mathbf{x}_i,\mathbf{x}_j) = \text{exp}\left(- \|\mathbf{x}_i - \mathbf{x}_j \|^2 / (2\sigma^2) \right)$. Input data was normalized to zero mean and unit variance, and the kernel width $\sigma$ was selected as the median of all pairwise distances between samples~\cite{Blaschko11}. The same $\sigma$ has been used for all methods, so that features are extracted from the same mapping of the input data. We can see that the non-linear mapping improves class separability. As expected, kPCA, kPLS and kCCA maximize in $\cal F$ the same variance, covariance and correlation objectives, respectively, as their linear counterparts. kOPLS looks for the directions of data in $\cal F$ that can provide the best approximation of $\Y$ in the MSE sense. This example illustrates also that maximizing the variance or even the covariance may not be the best choice for supervised learning.

Although kernel MVA methods can still be described in terms of linear equations, their direct solution faces several problems. In particular, it is well-known that kOPLS and kCCA can easily overfit the training data, so regularization is normally required to alleviate numerical instabilities~\cite{ShaweTaylor04,ArenasGarcia07}. A second important problem is related to the computational cost. Since $\Kbf_x$ is of size $l\times l$, methods' complexity scales quadratically with $l$ in terms of memory, and cubically with respect to the computation time. Further, the solution of the maximization problem (matrix $\A$) is not sparse, so that feature extraction for new data requires the evaluation of $l$ kernel functions {\em per} pattern, becoming computationally expensive for large $l$. Finally, it is worth mentioning the opposite situation: when $l$ is small, the extracted features may be useless, especially for high dimensional $\cal F$~\cite{Abrahansem11}. Actually, the information content of the features is elusive and has not been characterized so far. These issues limit the applicability of supervised kMVA in real-life scenarios with either very large or very small labeled data sets. In Section \ref{sec:sparsesemi}, we describe sparse and semisupervised approaches for kMVA that tackle both difficulties.

\subsection{Relations with other methods}

As already stated, close connections have been established among Fisher's LDA, CCA, PLS, and OPLS for classification~\cite{Barker03}; such links extend to their kernel counterparts as well. Under the framework of Rayleigh coefficients, Mika et al.~\cite{Mika99,Mika03} extended LDA to its kernel version for binary problems, and Baudat and Anouar~\cite{Baudat00} proposed the generalized discriminant analysis (GDA) for multiclass problems. A great many kernel discriminants have appeared since then, focused on alleviating problems such as those induced by high-dimensional small-sized datasets, the presence of noise, high levels of collinearity, or unbalanced classes. The number and heterogeneity of these methods makes difficult their unified treatment. 

Besides, in the recent years interesting connections of kMVA and statistical dependence estimates have been established. For instance, the Hilbert-Schmidt Independence Criterion (HSIC)~\cite{Gretton05COLT} is a simple yet very effective method to estimate statistical dependence between random variables. HSIC corresponds to estimating the norm of the cross-covariance in ${\mathcal F}$, whose empirical (biased) estimator is $\text{HSIC} := \frac{1}{(l-1)^2}\text{Tr}({\bf K}_x{\bf K}_y),$ where $\Kbf_x$ works with samples in the source domain and $\Kbf_y = \Y \Y^\top$. It can be shown that, if the RKHS kernel is universal, HSIC asymptotically tends to zero when the input and output data are independent. The so-called Hilbert-Schmidt Component Analysis (HSCA) method iteratively seeks for projections that maximize dependence with the target variables and simultaneously minimize the dependence with previously extracted features, both in HSIC terms. This objective translates into the iterative resolution of the generalized eigen-decomposition problem $\Kbf_x \Kbf_y \Kbf_x \aalpha = \lambda \Kbf_x \Kbf_f \Kbf_x \aalpha$, where $\Kbf_f$ is a kernel matrix of already extracted features in the previous iteration. If one is only interested in maximizing source-target dependence in HSIC terms, the problem boils down to solving kOPLS.

Similarly, the connection between kCCA and other kernel measures of dependence, such as the kernel Generalized Variance (kGV) or the kernel Mutual Information (kMI), was introduced in~\cite{Gretton05}. The empirical kGV estimates dependence between input-output data with a function that depends on the entire spectrum of the associated correlation operator in RKHS, $\text{kGV}(\theta)=-\frac{1}{2}\log(\displaystyle\mathop{\Pi}_i(1-\lambda_i^2))$, where $\lambda_i$ are the solutions to the generalized eigenvalue problem
\begin{equation}
\small
\begin{array}{ll} 
\left(\begin{array}{cc}
   \mathbf{0}       & \mathbf{K}_{x} \mathbf{K}_{y} \\ 
   \mathbf{K}_{y}\mathbf{K}_{x} & \mathbf{0}
\end{array}\right) 
\left(\begin{array}{c}
    \aalpha \\ 
    \vv 
\end{array}\right) = \\[5mm]
\lambda  
\left(\begin{array}{cc}
    \theta\mathbf{K}_{x}\mathbf{K}_{x}+\eta(1-\theta)\mathbf{K}_{x} & \mathbf{0}  \\ 
    \mathbf{0}                   & \theta\mathbf{K}_{y}\mathbf{K}_{y}+\eta(1-\theta)\mathbf{K}_{y} 
\end{array}\right) 
\left(\begin{array}{c}
    \aalpha \\ 
    \vv 
\end{array}\right), 
\end{array} \notag
\end{equation}
where $\mathbf{K}_{x}$ and $\mathbf{K}_{y}$ are defined using RKHS kernels obtained via convolution of the associated Parzen windows, $\eta$ is a scaling factor, and $\theta$ is a parameter in the range $[0,1]$. Gretton et al.~\cite{Gretton05} showed that, under certain conditions, kGV reduces to kMI for $\theta=0$ and to kCCA for $\theta=1$ (cf. Eq.~\ref{kernelitos}).

It is worth noting that the previous kernel measures of statistical dependence hold connections with Information Theoretic Learning concepts as well. For instance, it can be shown that HSIC is intimately related to the concept of correntropy~\cite{Principe10}. All these connections could shed light in the future about the informative content of the extracted features in a principled way.

\section{Extensions for large scale and semisupervised problems}
\label{sec:sparsesemi}

Supervised kernel MVA methods are hampered either by the wealth or the scarcity of labeled samples, which can make these methods impractical for many applications. We next summarize some extensions to deal with large scale problems and semisupervised situations in which few labeled data is available.

\subsection{Sparse Kernel Feature Extraction}

A critical bottleneck of kernel methods is that for a dataset of $l$ samples, the kernel matrices are $l\times l$, which, even for a moderate number of samples, quickly becomes a problem with respect to both memory and computation time. Furthermore, in kernel MVA this is also an issue during the extraction of features for test data, since kernel MVA solutions will in general depend on all training data (i.e., matrix $\A$ will generally be dense): evaluating thousands of kernels for every new input vector is, in most applications, simply not acceptable. Furthermore, these so-called {\em dense} solutions may result in severe problems of overfitting, which is particularly true for kOPLS and kCCA~\cite{ArenasGarcia07,ShaweTaylor04}. To address these problems, several solutions have been proposed to obtain sparse solutions that can be expressed as a combination of a subset of the training data, and therefore require only $r$ kernel evaluations {\em per} sample (with $r \ll l$) for feature extraction. Note that, in contrast to the many linear MVA algorithms that induce sparsity with respect to the original variables, in this subsection we review methods that obtain sparse solutions in terms of the samples (i.e., sparsity in the $\aalpha_i$ vectors).

The approaches to obtain sparse solutions can be broadly divided into {\em low rank} approximation methods, that aim at working with reduced $r\times r$ matrices ($r\ll l$), and {\em reduced set} methods that work with $l\times r$ matrices. Following the first approach, the Nystr\"om low-rank approximation of an $l\times l$ kernel matrix ${\bf K}_{ll}$ is expressed as $\tilde{\bf K}_{ll} = {\bf K}_{lr}{\bf K}_{rr}^{-1}{\bf K}_{rl}$, where subscripts indicate row and column dimensions. The method was originally exploited in the context of Gaussian processes, and was later used in~\cite{hoegaerts2005} to directly approximate the feature mapping itself rather than the kernel, thus giving rise to sparse versions of kPLS and kCCA.

Among the reduced set methods, a sparse kPCA (skPCA) was proposed by Tipping in~\cite{Tipping01}, where the sparsity in the representation is obtained by assuming a generative model for the data in $\cal F$ that follows a normal distribution and includes a noise term with variance $v_n$. The maximum likelihood estimation of the covariance matrix is shown to depend on just a subset of the training data, and so it does the resulting solution. A sparse kPLS (skPLS) was introduced in~\cite{Bennett03}. The method uses a fraction of the training samples for computing the projections. Each projection vector is found through a sort of $\varepsilon$-insensitive loss similar to the one used in the support vector regression method. The sparsification is induced via a multi-step adaptation with high computational burden. In spite of obtaining sparse solutions, the algorithms from~\cite{Tipping01} and~\cite{Bennett03} still require the computation of the full kernel matrix during the training.

A reduced complexity kOPLS (rkOPLS) was proposed in~\cite{ArenasGarcia07} by imposing sparsity in the projection vectors representation {\em a priori}, $\U = \PHI_r^\top\BETA$, where $\PHI_r$ is a subset of the training data containing $r$ samples ($r \ll l$) and $\BETA$ is the new argument for the maximization problem, which now becomes: %. The problem now becomes:
\begin{equation}
\begin{aligned}
 \max & ~~ \BETA^\top \Kbf_{rl} \Y \Y^\top \Kbf_{rl}^\top \BETA \\
{\rm subject~to:} & ~~ \BETA^\top \Kbf_{rl} \Kbf_{rl}^\top \BETA = 1, \\
\end{aligned}
%\vspace{-.1cm}
\end{equation}
Since kernel matrix $\Kbf_{rl} = \PHI_r \PHI^\top$ involves the inner products in $\cal F$ of all training points with the patterns in the reduced set, this method still takes into account all data during the training phase, and is therefore different from simple subsampling. This sparsification procedure avoids the computation of the full kernel matrix at any step of the algorithm. An additional advantage of this method is that matrices $\Kbf_{rl} \Y \Y^\top \Kbf_{rl}^\top$ and $\Kbf_{rl} \Kbf_{rl}^\top$ are both of size $r \times r$, and can be expressed as sums over the training data, so the storage requirements become just quadratic with $r$. Furthermore, the sparsity constraint acts as a regularizer that can significantly improve the generalization ability of the method. In the experiments section, we will apply the same sparsification procedure to kPCA and kCCA, obtaining reduced complexity versions of these methods to which we will refer to as rkPCA and rkCCA. Interestingly, the extension to kPLS2 is not straightforward, since the deflation step would still require the full kernel matrix $\mathbf{K}_{ll}$.

Alternatively, two sparse kPLS schemes were presented in~\cite{Dhanjal09} under the name of Sparse Maximal Aligment (SMA) and Sparse Maximal Covariance (SMC). Here kPLS iteratively estimates projections that either maximize the {\em kernel alignment} (c.1) or the {\em covariance} (c.2) of the projected data and the true labels:
\begin{equation}
\begin{aligned}
 \max & ~~ \BETA^\top \Kbf_j \Y \\
{\rm subject~to~(c.1):} & ~~ \BETA^\top \Kbf_j^2 \BETA = 1, \\
{\rm subject~to~(c.2):} & ~~ \BETA^\top \Kbf_j\BETA = 1, \\
%     {\rm subject~to:}             & ~~ \text{card}(\BETA)=1, \\
\end{aligned}
%\vspace{-.1cm}
\end{equation}
where $\Kbf_1 = \Kbf_x$ and $\Kbf_j$ denotes the deflated kernel matrix at iteration $j$, according to \cite[Eq. (3)]{Dhanjal09}. The method imposes the additional constraint that the cardinality of $\BETA$ is 1. This restriction explicitly enforces sparsity through an $\ell_0$-norm in the weights space. At each iteration, $\BETA$ is obtained by performing an exhaustive search over all training patterns. However, the complexity of the algorithm can be significantly reduced by constraining the search to just $p$ randomly chosen samples.

\begin{table*}[t!]
\begin{center}
\caption{Main properties of (k)MVA methods. Computational complexity and implementation issues are categorized for the considered dense and sparse methods, indicating from left to right: the free parameters, number of kernel evaluations (KE) during training, storage requirements, whether an eigenproblem (EIG) or generalized eigenproblem (GEV) needs to be solved, and the number of kernels that need to be evaluated to extract projections for new data.\label{tab_summary}}
\begin{tabular}{l|ccccc}
\hline
\hline
Method & Parameters & KE tr. & Storage Req. & EIG / GEV / Other & KE test \\
\hline
PCA & none & none & ${\cal O}(d^2)$ & EIG & none\\
PLS & none & none & ${\cal O}((d+m)^2)$ & EIG & none\\
CCA & none & none & ${\cal O}((d+m)^2)$ & GEV & none\\
OPLS & none & none & ${\cal O}(d^2)$ & GEV & none\\
\hline
kPCA & kernel & $l^2$ & ${\cal O}(l^2)$ & EIG & $l$\\
kPLS & kernel & $l^2$ & ${\cal O}((l+m)^2)$ & EIG & $l$\\
kCCA & kernel & $l^2$ & ${\cal O}((l+m)^2)$ & GEV & $l$\\
kOPLS & kernel & $l^2$ & ${\cal O}(l^2)$ & GEV & $l$ \\
\hline
skPCA~\cite{Tipping01} & kernel, $v_n$ & $l^2$ & ${\cal O}(l^2)$ & ML + EIG$^\dagger$ & $v_n$ dependent\\
skPLS~\cite{Bennett03} & kernel, $\nu$, $\varepsilon$ & $l^2$ & ${\cal O}(l^2)$ & $\nu$-SVR& $\nu, \varepsilon$ dependent\\
rkPCA & kernel, $r$ & $r l$ & ${\cal O}(r^2)$ & EIG & $r$ \\
rkCCA & kernel, $r$ & $r l$ & ${\cal O}((r+m)^2)$ & GEV & $r$ \\
rkOPLS~\cite{ArenasGarcia07}& kernel, $r$ & $r l$ & ${\cal O}(r^2)$ & GEV & $r$ \\
SMA / SMC~\cite{Dhanjal09}$^\ddagger$ & kernel, $r$ & $l^2$ & ${\cal O}(l^2)$ & Ex. search & $r$\\
\hline
\hline
\end{tabular}
\end{center}
\footnotesize{$^\dagger$ A maximum likelihood estimation step is required. $^\ddagger$ By constraining the search to $p$ random samples at each step of the algorithm, kernel evaluations and storage requirements during training can be reduced to $r l p$ and ${\cal O}(l p)$, respectively.}
\end{table*}

In Table~\ref{tab_summary}, we summarize some computational and implementation issues of the aforementioned sparse kMVA methods, and of standard non-sparse kMVA and linear methods. An analysis of the properties of each algorithm provides some hints that can help us choose the algorithm for a particular application. Firstly, a critical step when using kernel methods is the selection of an appropriate kernel function and tuning its parameters. To avoid overfitting, kMVA methods can be adjusted using cross-validation at the cost of higher computational cost. Sparse methods can help in this respect by regularizing the solution. Secondly, most methods can be implemented as either eigenvalue or generalized eigenvalue problems, whose complexity typically scales cubically with the size of the analyzed matrices. Therefore, both for memory and computational reasons, only linear MVA and the sparse approaches from~\cite{ArenasGarcia07} and~\cite{Dhanjal09} are affordable when dealing with large data sets. A final advantage of sparse kMVA is the reduced number of kernel evaluations to extract features for new data.

\subsection{Semisupervised Kernel Feature Extraction}

When few labeled samples are available, the extracted features do not capture the structure of the data manifold well, and hence using them for classification or regression may lead to very poor results. Recently, semisupervised learning approaches have been introduced to alleviate these problems. Two approaches are encountered: the information conveyed by the unlabeled samples is either modeled with graphs or via kernel functions derived from generative clustering models.

Notationally, we are given $l$ labeled and $u$ unlabeled samples, a total of $n=l+u$. The semisupervised kCCA (ss-kCCA) has been recently introduced in~\cite{Blaschko11} by using the graph Laplacian. The method essentially solves the standard kCCA using kernel matrices computed with both labeled and unlabeled data, which are further {\em regularized} with the graph Laplacian:
\begin{eqnarray}
 &\left( \begin{array}{cc} \bf 0 & \Kbf_{nl}^x \Kbf_{ln}^y \\ \Kbf_{nl}^x \Kbf_{ln}^y & \bf0 \end{array}  \right) \left( \begin{array}{c} \aalpha \\ \vv \end{array}  \right) = \nonumber\\[5mm]
& \lambda \left( \begin{array}{cc} \Kbf_{nl}^x\Kbf_{ln}^x + \Rbf_{nn}^x & \bf0 \\  \bf0 & \Kbf_{nl}^y \Kbf_{ln}^y + \Rbf_{nn}^y \end{array}  \right) \left( \begin{array}{c} \aalpha \\ \vv \end{array}  \right),
\end{eqnarray}
where $\Rbf_{nn}^x = \alpha_x\Kbf_{nn}^x + \gamma_x\Kbf_{nn}^x\boldsymbol{\mathcal L}_{nn}^x\Kbf_{nn}^x$ and $\Rbf_{nn}^y = \alpha_y\Kbf_{nn}^y + \gamma_y\Kbf_{nn}^y\boldsymbol{\mathcal L}_{nn}^y\Kbf_{nn}^y$. For notation compactness, subindexes here indicate the size of the corresponding matrices while superscripts denote whether they involve input or output data. Parameters $\alpha_x$, $\alpha_y$, $\gamma_x$ and $\gamma_y$ trade off the contribution of labeled and unlabeled samples, and $\boldsymbol{\mathcal L}={\bf D}^{-1/2}({\bf D}-{\bf M}){\bf D}^{1/2}$ represents the (normalized) graph Laplacian for the input and target domains, where ${\bf D}$ is the degree matrix whose entries are the sums of the rows of the corresponding similarity matrix ${\bf M}$, i.e. ${\bf D}_{ii}=\sum_j {\bf M}_{ij}$. It should be noted that for $n=l$ and null regularization, one obtains the standard kCCA (cf. Eq.~\ref{kernelitos}). Note also that this form of regularization through the graph Laplacian can be applied to any kMVA method. A drawback of this approach is that it involves tuning several parameters and working with larger matrices of size $2n\times 2n$, which can make its application difficult in practice.

Alternatively, cluster kernels have been used to develop semisupervised versions of kernel methods in general, and of kMVA methods in particular. The approach was used for kPLS and kOPLS in~\cite{Izquierdo12}. Essentially, the method relies on combining a kernel function based on labeled information only, $k_s({\bf x}_i,{\bf x}_j)$, and a {\em generative} kernel directly learned by clustering {\em all} (labeled and unlabeled) data, $k_c({\bf x}_i,{\bf x}_j)$. Building $k_c$ requires first running a clustering algorithm, such as Expectation-Maximization assuming a Gaussian mixture model (GMM) with different initializations, $q=1,\ldots,Q$, and with different number of clusters, $g=2,\ldots,G+1$. This results in $Q\cdot G$ cluster assignments where each sample ${\bf x}_i$ has its corresponding posterior probability vector $\boldsymbol{\pi}_i(q,g)\in\Real^g$. The {\em probabilistic cluster} kernel $k_{c}$ is computed by averaging all the dot products between posterior probability vectors, $$k_c(\x_i,\x_j) = \frac{1}{Z}\sum_{q=1}^Q\sum_{g=2}^{G+1} \boldsymbol{\pi}_i(q,g)^\top\boldsymbol{\pi}_j(q,g),$$ 
where $Z$ is a normalization factor. The final kernel function is defined as the weighted sum of both kernels, $k({\bf x}_i,{\bf x}_j) = \beta k_{s}({\bf x}_i,{\bf x}_j) + (1-\beta) k_{c}({\bf x}_i,{\bf x}_j)$,
where $\beta\in[0,1]$ is a scalar parameter to be adjusted. Intuitively, the cluster kernel accounts for {\em probabilistic} similarities at small and large scales (number of clusters) between all samples along the data manifold. The method does not require computationally demanding procedures (e.g. current GMM clustering algorithms scale linearly with $n$), and the kMVA still relies just on the labeled data, and thus requires an $l\times l$ kernel matrix. All these properties are quite appealing from the practitioner's point of view.

\section{Experimental Results}\label{sec:experiments}

In this section, we illustrate through different application examples the use and capabilities of the supervised kernel multivariate feature extraction framework. We start by comparing the performance of the linear and kernel MVA methods in a benchmark of classification problems from the publicly available Machine Learning Repository at University of California, Irvine (UCI)\footnote{http://archive.ics.uci.edu/ml/.}.  We then consider two real applications to show the potential of these algorithms: satellite image processing~\cite{ArenasGarcia08} and audio processing for music genre prediction~\cite{Meng07}. The size of the data set used in this second scenario is sufficiently large to make standard kMVA methods impractical, a situation that we will use to illustrate the benefits of the sparse extensions.

\subsection{UCI repository benchmark}

Our first battery of experiments deals with standard benchmark data sets taken from the UCI repository, and will be oriented to discuss some important properties of the standard linear and kernel MVA methods. The main properties of the selected problems are given in Table \ref{uciresults}, namely the number of training and test patterns ($l$, $l_\text{test}$), the dimensionality of the input space ($d$), the number of classes ($c$), the ratio of training patterns {\em per} dimension, and the Kullback-Leibler divergence (KL) between the sample probabilities of each class and a uniform distribution, that can be seen as an indicator of balance among classes. The train/test partition has left $60\%$ of the total data (or alternatively a maximum of $500$ samples) in the training set, so that standard kMVA complexity is kept under control. Since all selected problems involve a classification task, matrix $\Y$ was used to store the class information using 1-of-$c$ coding. To obtain the overall accuracies (OA) displayed in the table, we have trained an LS model to approximate $\Y$, followed by a ``winner-takes-all'' activation function.

We start the discussion by comparing the performance of linear methods. For OPLS and CCA we have used the maximum number of features ($c - 1$), whereas for PCA $n_f$ has been fixed through a 10-fold cross-validation scheme on the training set, and is indicated next to the results of the method. When the maximum number of features are extracted, linear OPLS and CCA become identical. We can see that in most cases they perform similarly to PCA, but requiring significantly fewer features. It is important to see the very poor performance of OPLS and CCA in two particular problems: {\em semeion} and {\em sonar}. We see that the ratio $l/d$ is very small for these problems, so the input covariance matrix is likely to be ill-conditioned. Then, very low variance directions of the input space are used by CCA and OPLS to overfit the data. To avoid this problem, it becomes necessary to regularize these algorithms, e.g., by loading the main diagonal of the covariance matrix $\C_x$ or by executing the method on the projections already extracted by PCA~\cite{Braun08}. The results of regularized OPLS and CCA following the latter approach, and using the maximum of $(c-1)$ features, are given in the last two columns of Table~\ref{uciresults}, where we can see that the regularization does indeed help to overcome the ``small-sample-size'' problem. For completeness, we present also the results of the PLS2 approach. To get a fair comparison with the other supervised schemes, PLS2 is also trained to extract $(c-1)$ features. Thus, we can conclude that OPLS and CCA allow to obtain more discriminative features than the other methods, but one needs to be aware of the likely need to regularize the solution.

Next, we turn our attention to non-linear versions, whose results have been displayed in the bottom half of the table. An RBF kernel has been used in all cases, selecting the kernel width with 10-fold cross-validation. A first consideration is that kernel approaches considerably outperform the linear schemes. Since the RBF kernel implies a mapping to a very high dimensional space, it is not surprising that standard kOPLS and kCCA are even more prone to overfitting than before, this being a well-known problem of these methods. As before, regularized solutions allow kOPLS and kCCA to achieve comparable performance to kPCA, but retaining a much smaller number of features $(c-1)$, which demonstrates the superior discriminative capabilities of the features extracted by these methods. For completeness, kPLS2 was used to extract the same number of features as kOPLS and kCCA, achieving considerably smaller OAs in all problems. Nevertheless, with kPLS2 it is possible to  extract a larger number of projections, and this method is known to be more robust to overfitting than kOPLS and kCCA.

\newcolumntype{g}{>{\columncolor{Gray}}c}
\begin{table*}[t!]
\begin{center}
\caption{Main characteristics of the data sets that compose the UCI benchmark and performance of the different (k)MVA feature extraction methods. As a figure of merit we use the overall accuracy (OA, [\%]) $\pm$ the binomial standard deviation. Best results for each problem are highlighted in boldface. The number of extracted features is indicated for PCA and kPCA, whereas all other methods use $c-1$ features.\label{uciresults}}
\footnotesize
\setlength{\tabcolsep}{2pt}
\begin{small}
\begin{tabular}{l|@{\hskip .15cm}ccc|crl@{\hskip .4cm}rl@{\hskip .4cm}rl@{\hskip .4cm}rl@{\hskip .4cm}rl@{\hskip .4cm}rl}
\hline
\hline
data set &~$l$~&~$d$~~&~$l/d$~~& $n_{f,\text{PCA}}$ & \multicolumn{2}{c@{\hskip .4cm}}{PCA} & \multicolumn{2}{c@{\hskip .4cm}}{PLS2} & \multicolumn{2}{c@{\hskip .4cm}}{OPLS} & \multicolumn{2}{c@{\hskip .4cm}}{CCA} & \multicolumn{2}{c@{\hskip .4cm}}{reg. OPLS} & \multicolumn{2}{c@{\hskip .4cm}}{reg. CCA} \\
\hline
car & 500 & 6 & 83.3 & 5 &79 &$\pm$1.2 &79.3 &$\pm$1.2 &79.6 &$\pm$1.1 &79.6 &$\pm$1.1 &79 &$\pm$1.2 &79 &$\pm$1.2 \\
glass & 128 & 9 & 14.2 & 9 &57 &$\pm$5.3 &50 &$\pm$5.4 &57 &$\pm$5.3 &57 &$\pm$5.3 &57 &$\pm$5.3 &57 &$\pm$5.3 \\
optdigits & 500 & 62 & 8.06 & 27 &88.8 &$\pm$0.4 &86.6 &$\pm$0.5 &90.3 &$\pm$0.4 &90.3 &$\pm$0.4 &88.8 &$\pm$0.4  &88.8 &$\pm$0.4 \\ 
semeion & 500 & 256 & 1.95 & 78 & 83.8 &$\pm$1.1 &82.4 &$\pm$1.2 &69.1 &$\pm$1.4 &69.1 &$\pm$1.4 &83.8 &$\pm$1.1 &83.8 &$\pm$1.1 \\ 
sonar & 125 & 60 & 2.08 & 7 &74.7 &$\pm$4.8 &67.5 &$\pm$5.1 &65.1 &$\pm$5.2 &65.1 &$\pm$5.2 &74.7 &$\pm$4.8 &74.7 &$\pm$4.8 \\
vehicle & 500 & 18 & 27.8 & 18 	&78 &$\pm$2.2 &63.9 &$\pm$2.6 &78  &$\pm$2.2 &78 &$\pm$2.2 &78 &$\pm$2.2 &78  &$\pm$2.2 \\ 
vowel & 500 & 13 & 38.5 & 13 &48.8 &$\pm$2.3 &46.9 &$\pm$2.3 &48.8 &$\pm$2.3 &48.8 &$\pm$2.3 &48.8 &$\pm$2.3 &48.8 &$\pm$2.3 \\
yeast & 500 & 8 & 62.5 & 7 &55.9 &$\pm$1.6 &55 &$\pm$1.6 &55 &$\pm$1.6 &55 &$\pm$1.6 &55.9 &$\pm$1.6 &55.9 &$\pm$1.6 \\
\hline
data set &~$l_\text{test}$~&~$c$~~&~KL~~& $n_{f,\text{kPCA}}$ & \multicolumn{2}{c@{\hskip .4cm}}{kPCA} & \multicolumn{2}{c@{\hskip .4cm}}{kPLS2} & \multicolumn{2}{c@{\hskip .4cm}}{kOPLS} & \multicolumn{2}{c@{\hskip .4cm}}{kCCA} & \multicolumn{2}{c@{\hskip .4cm}}{reg. kOPLS} & \multicolumn{2}{c@{\hskip .4cm}}{reg. kCCA} \\
\hline
car & 1228 & 4 & 0.55 & 197 &{\bf 93} &$\pm$0.7 &80.8 &$\pm$1.1 &92.1 &$\pm$0.8 &71.8 &$\pm$1.3 &{\bf 93} &$\pm$0.7 &91.6 &$\pm$0.8 \\
glass & 86 & 6 & 0.28 & 17 &60.5 &$\pm$5.3 &60.5 &$\pm$5.3 &41.9 &$\pm$5.3 &29.1 &$\pm$4.9 &{\bf 62.8} &$\pm$5.2 &60.5 &$\pm$5.3 \\
optdigits & 5120 & 10 & 0 & 330 &{\bf 95.4} &$\pm$0.3  &82.2 &$\pm$0.5 &95.2 &$\pm$0.3 &93.3 &$\pm$0.4 &95.3 &$\pm$0.3 &88.8 &$\pm$0.4 \\ 
semeion & 1093 & 10 & 0 & 321 &{\bf 89.5} &$\pm$0.9 &79 &$\pm$1.2 &89.3 &$\pm$0.9 &89.4 &$\pm$0.9 &89 &$\pm$0.9 &83.9 &$\pm$1.1 \\ 
sonar & 83 & 2 & 0 & 97 &{\bf 84.3} &$\pm$4 &67.5 &$\pm$5.1 &80.7 &$\pm$4.3 &80.7 &$\pm$4.3 &{\bf 84.3} &$\pm$4 &49.4 &$\pm$5.5\\
vehicle & 346 & 4 & 0 & 229 	&81.5 &$\pm$2.1 &49.7 &$\pm$2.7 &76.6 &$\pm$2.3 &75.1 &$\pm$2.3 &{\bf 82.1} &$\pm$2.1 &72.8 &$\pm$2.4 \\ 
vowel & 490 & 11 & 0 & 310 &92.7 &$\pm$1.2 &53.1 &$\pm$2.3 &92.4 &$\pm$1.2 &92 &$\pm$1.2 &{\bf 93.1} &$\pm$1.1 &88.4 &$\pm$1.4 \\
yeast & 984 & 10 & 0.58 & 56 &58.8 &$\pm$1.6 &56.8 &$\pm$1.6 &48.1 &$\pm$1.6 &33.8 &$\pm$1.5 & 58.7 &$\pm$1.6 & {\bf 58.9} &$\pm$1.6 \\
\hline
\hline
\end{tabular}
\end{small}
\vspace{-.5cm}
\end{center}
\end{table*}

\subsection{Remote sensing image analysis}

The last few hundred years human activities have precipitated an environmental crisis on Earth. %Quantification of the impact on the Earth's system are matter of current and intense research.
In the last decade, advanced statistical methods have been introduced to quantify our impact on the land/vegetation and  atmosphere, to better understand their interactions. Nowadays, multi- and hyper-spectral sensors mounted on satellite or airborne platforms may acquire the reflected energy by the Earth with high spatial detail and in several wavelengths. Recent infrared sounders also allow us to estimate the profiles of atmospheric parameters with unprecedented accuracy and vertical resolution. Here, we pay attention to the performance of several kMVA methods for both image segmentation of hyperspectral images, and estimation of climate parameters from infrared sounders.

\paragraph{Hyperspectral image classification} The first case study deals with image segmentation of hyperspectral images~\cite{ArenasGarcia09}. We have used the standard AVIRIS image taken over NW Indiana's Indian Pine test site in June 1992\footnote{The calibrated data is available online (along with  detailed ground-truth information) from http://dynamo.ecn.purdue.edu/~biehl/MultiSpec.}. We removed $20$ noisy bands covering the region of water absorption, and finally worked with $200$ spectral bands. The high number of narrow spectral bands induce a high collinearity among features. Discriminating among the major crop classes in the area can be very difficult (in particular, given the moderate spatial resolution of 20 meters), which has made the scene a  challenging benchmark to validate classification accuracy of hyperspectral imaging algorithms. The image is $145\times 145$ pixels and contains $16$ quite unbalanced classes (ranging from $20-2468$ pixels). Among the available $10366$ labeled pixels, $20\%$ were used for training the feature extractors, and the remaining $80\%$ for testing. The discriminative power of all extracted features was tested using a simple classifier consisting of a linear least squares model followed by a ``winner-takes all'' activation function.

\begin{figure*}[t!]
\centerline{\IG[width=14cm]{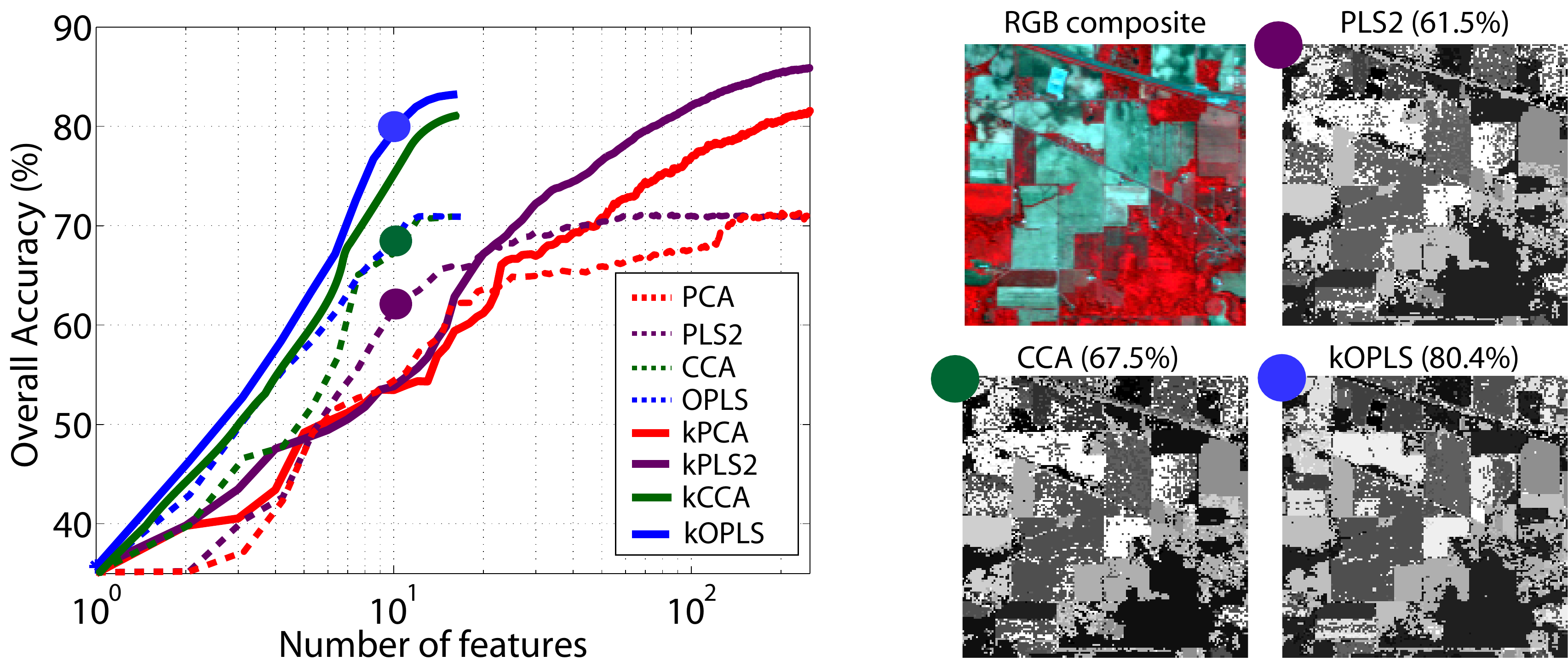}}
\caption{Average classification accuracy (\%) for linear and kernel MVA methods as a function of the number of extracted features, along some classification maps for the case of $10$ extracted features.}
\label{fig_aviris}
\end{figure*}

Figure~\ref{fig_aviris} shows the test classification accuracy for a varying number of extracted features, $n_f$. For linear models, OPLS performs better than all other methods for any number of extracted features. Even though CCA provides similar results for $n_f=10$, it involves a slightly more complex generalized eigenproblem. When the maximum number of projections is used, all methods result in the same error. Nevertheless, while PCA and PLS2 require $200$ features (i.e., the dimensionality of the input space), CCA and OPLS only need $15$ features to achieve virtually the same performance.

We also considered non-linear kPCA, kPLS2, kOPLS and kCCA, using an RBF kernel whose width was adjusted using $5$-fold cross-validation in the training set. The same conclusions obtained for the linear case apply also to MVA methods in kernel feature space. The features extracted by kOPLS allow to achieve a slightly better Overall Accuracy (OA) than kCCA, and both methods perform significantly better than kPLS2 and kPCA. In the limit of $n_f$, all methods achieve similar accuracy. The classification maps obtained for $n_f= 10$ confirm these conclusions: higher accuracies lead to smoother maps and smaller error in large spatially homogeneous vegetation covers.

\paragraph{Temperature estimation from infrared sounding data} The second case study focuses on the estimation of temperature from spaceborne very high spectral resolution infrared sounding instruments. Despite the constant advances in sensor design and retrieval techniques, it is not trivial to invert the full information of the atmospheric state contained by such hyperspectral measurements. Statistical regression and feature extraction methods have overcome the numerical difficulties of radiative transfer models (RTMs), and enable fast retrievals from high volumes of data~\cite{CampsValls12}.

We concentrate here on the Infrared Atmospheric Sounding Interferometer (IASI) onboard the MetOp-A satellite data. IASI spectra consist of $8461$ spectral channels (input features) with a spatial resolution of $25$~km at nadir. Due to its large spatial coverage and low radiometric noise, IASI provides twice daily global measurements of key atmospheric species such as ozone, carbon monoxide, methane and methanol. Due to the impossibility to obtain real radiosound data for the whole atmospheric column, we resorted to the standard hybrid approach for developing the prediction models: we used synthetic data for training the models, and then applied it to a full IASI orbit (91800 `pixels' on March 4th, 2008). A total amount of 67475 synthetic samples were simulated with an infrared RTM according to input profiles of temperature at $90$ pressure levels. 

\begin{figure*}[t!]
\centerline{\IG[width=17cm]{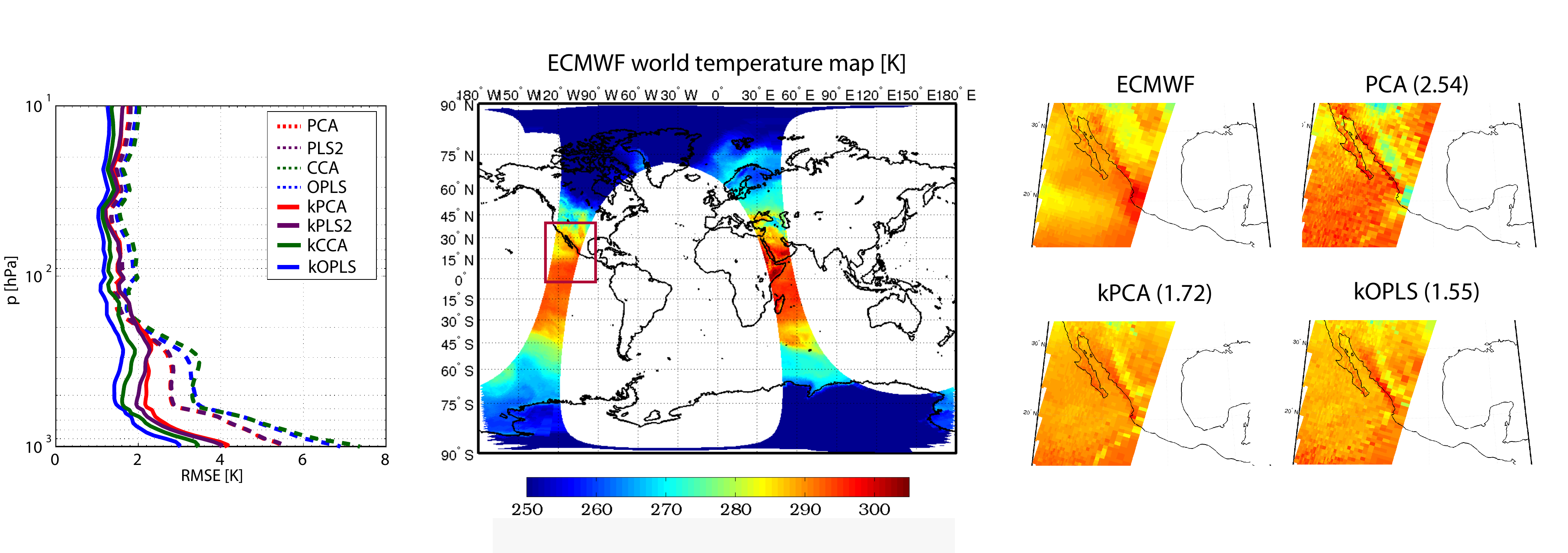}}
\caption{RMSE atmospheric temperature profiles (left); surface temperature [K] world map provided by the official ECMWF model, http://www.ecmwf.int/, on March 4, 2008 (middle); and estimated surface temperature maps in California/Mexico area for several methods along with the averaged RMSE across the whole atmospheric column given in brackets.}
\label{fig_iasi}
\end{figure*}

We are confronted here with a challenging multi-output regression problem ($\x_i\in\Real^{8461}$ and $\y_i\in\Real^{90}$), which needs fast methods for retrieval (prediction). Note that the IASI mission delivers approximately $1.3\times10^6$ spectra per day, which gives a rate of about 29 Gbytes/day to be processed. We compared several MVA methods followed by LS in terms of the root-MSE (RMSE) computed as the discrepancy to the official European Centre for Medium-Range Weather Forecasts (ECMWF) estimations. We obtain the RMSE for all spatial `pixels' in the orbit and for all layers of the atmosphere. We fixed a maximum $n_f=100$, and used only $l=2000$ samples for learning the transformation and the LS regression weights. For the kernel approaches, an RBF kernel is used, selecting the width using 5-fold cross validation.

The left panel of Fig.~\ref{fig_iasi} illustrates the RMSE obtained for different pressure levels, using a spatial average over all the pixels. Results show that kernel methods outperform linear approaches in RMSE terms, specially in the lower parts of the atmosphere, probably due to the presence of clouds and haze. Estimated surface temperature maps for a particular area are also given in the right panel of Fig.~\ref{fig_iasi} for PCA, kPCA, and kOPLS. These plots reveal that kernel methods yield closer maps to those provided by the ECMWF in averaged RMSE terms. These results can be of high value because the kernel-based estimations are obtained with a drastic reduction in computational time compared to the physical-inversion methods used by the ECMWF.

\subsection{Analysis of integrated short-time music features for genre prediction}
In this subsection, we consider the problem of predicting the genre of a song using the audio data only, a task which has recently been a subject of much interest.  The data set we analyze has been previously used in~\cite{Meng07,ArenasGarcia07}, and consists of 1317 snippets of 30 seconds distributed evenly over $11$ music genres: alternative, country, easy listening, electronica, jazz, latin, pop\&dance, rap\&hip-hop, r\&b, reggae and rock. This is a rather complex data set with an average of $1.83$ songs per artist. An estimate of human performance on this data set has been carried out via subjective tests, providing an average accuracy rate around $55\%$.

The music snippets are MP3 (MPEG1-layer3) encoded music with a bitrate of 128 kbps or higher, down sampled to 22050 Hz, and they are processed following the method in~\cite{Meng07}: Mel Frequency Cepstral Coefficients (MFCC) features are extracted from overlapping frames of the song, using a window size of 20 ms. Then, a multivariate autoregressive (AR) model is adjusted for every 1.2 seconds of the song to capture temporal correlation, and finally the parameters of the AR model are stacked into a 135 length feature vector for every such frame.

For training and testing the system we have split the data set into two subsets with 817 and 500 songs, respectively. After processing the audio data, we have $57388$ and $36556$ 135-dimensional vectors in the training and test partitions, an amount which for standard kernel MVA methods is prohibitively large. For this reason, in this subsection we study the performance of linear MVA methods, as well as of sparse kernel methods that promote sparsity following the approach of~\cite{ArenasGarcia07}: rkPCA, rkOPLS, and rkCCA. For completeness, we have also considered the kPLS2 method of~\cite{ShaweTaylor04}; in this case the deflation scheme does not allow to use reduced set methods, so we applied mere subsampling. As in the previous subsections, we use an RBF kernel, with a 10-fold validation scheme over the training data to adjust the kernel width. We also resorted to an LS scheme followed by ``winner-takes-all'' to carry out the classification.

\begin{figure*}[t!]
\centerline{\IG[width=8cm]{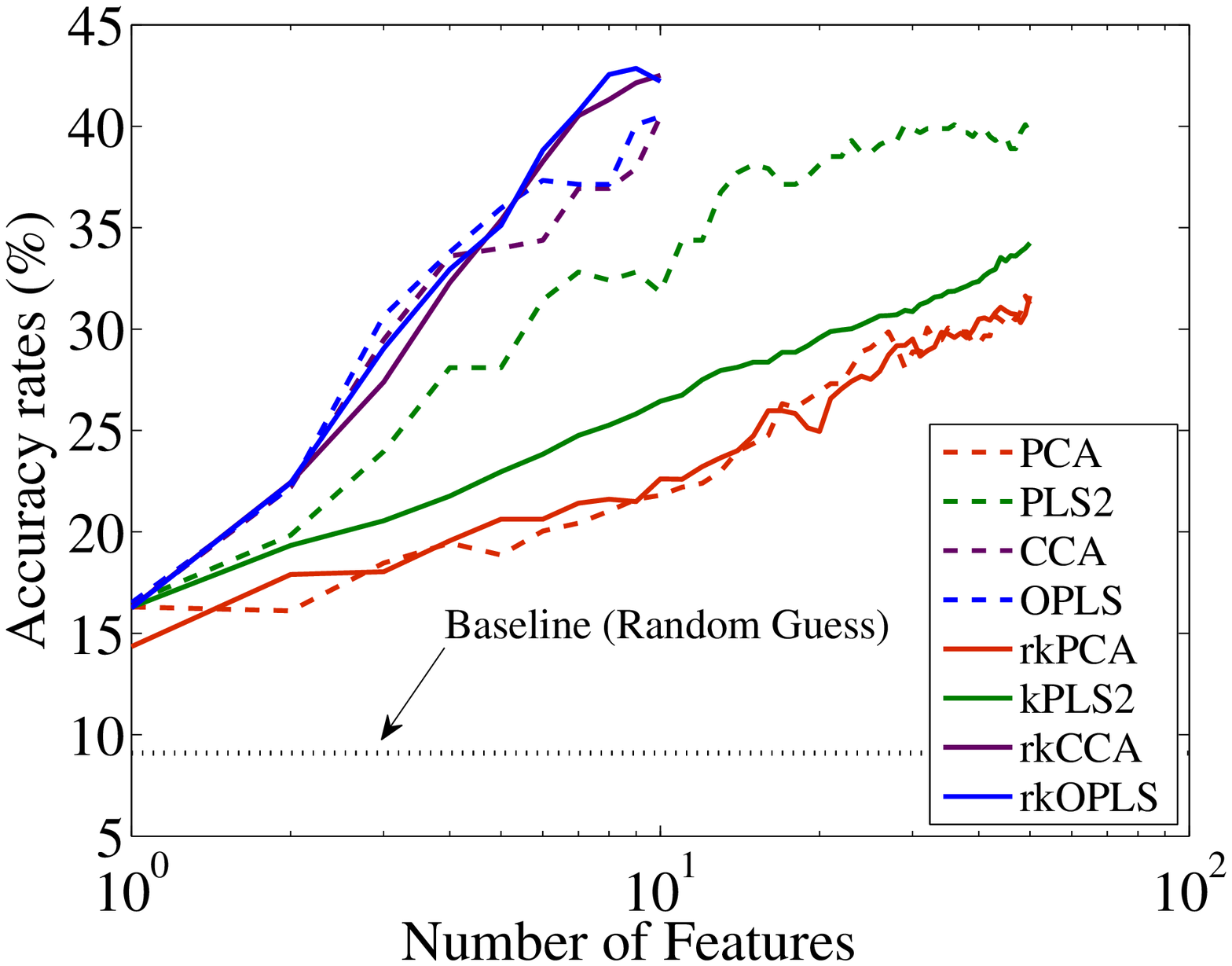} \IG[width=8cm]{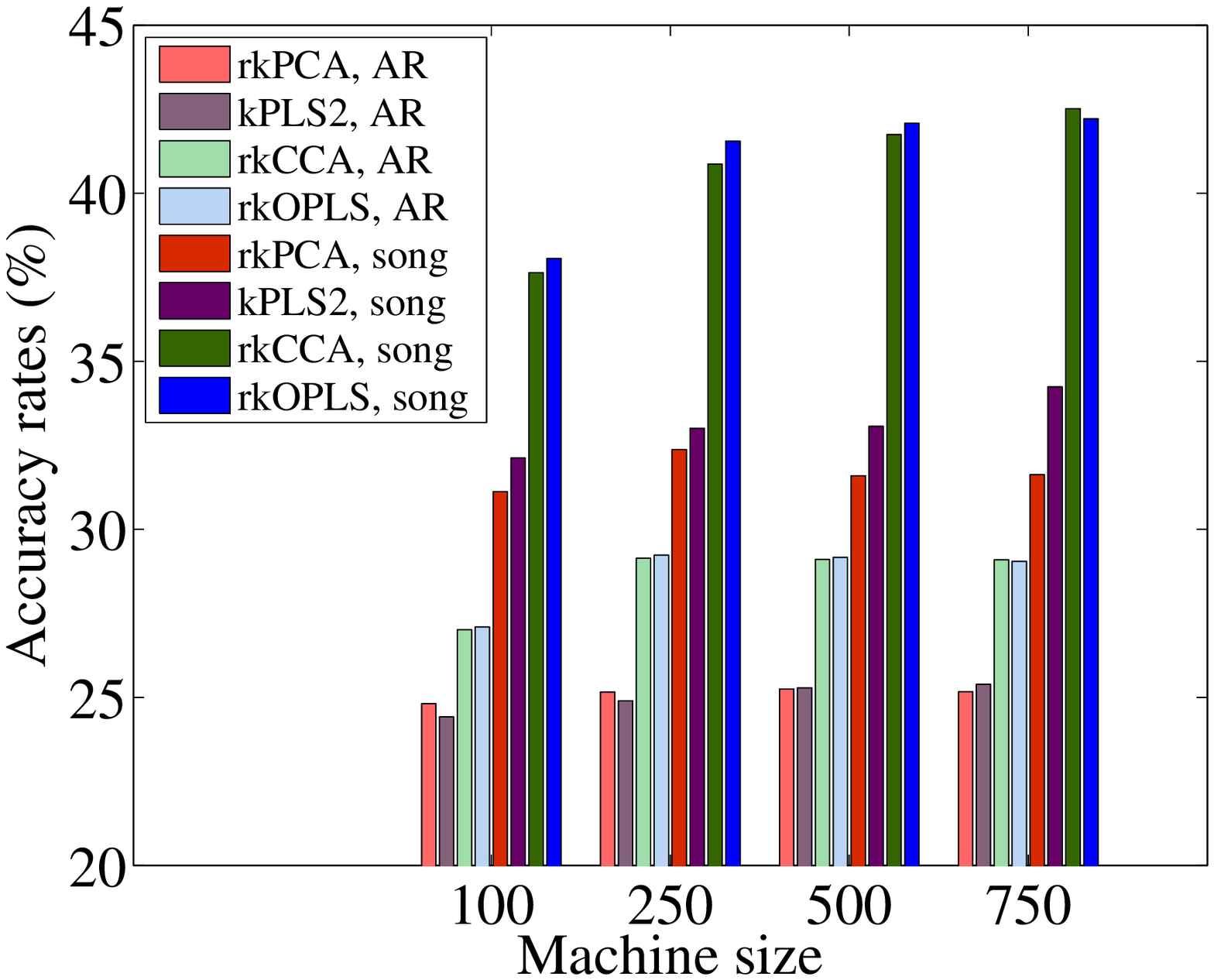}}
\caption{Genre classification accuracy of linear and sparse kernel MVA methods. Left: Accuracy at the song level as a function of the number of extracted features. Right: Accuracy of the sparse methods measured as percentage of correctly classified AR patterns and songs for different machine sizes.}
\label{fig_genre1}
\end{figure*}

Figure~\ref{fig_genre1} illustrates the performance of all methods for different number of features. For the sparse methods, the influence of the machine size, measured as the number of kernel evaluations required to extract features for new data, is also analyzed. In rkPCA, rkOPLS and rkCCA this is given by the cardinality of the reduced set, whereas in kPLS2 it coincides with the number of samples used to train the feature extractor. Since every song consists of about $70$ AR vectors, we can measure the classification accuracy in two different ways: 1) On the level of individual AR vectors, or 2) by majority voting across the AR vectors of a given song. The left panel of Fig.~\ref{fig_genre1} illustrates the discriminative power of the features extracted by all considered methods. Overall, the best performance is obtained by OPLS- and CCA-type methods, with the kernel schemes outperforming the linear ones for $n_f > 5$. The poor performance of PCA and rkPCA makes evident the need of exploiting label information during the feature extraction step. Finally, it is evident that a mere subsampling does not provide kPLS2 with enough data to extract relevant features, and this method performs worse than its linear counterpart.

On the right panel of Fig.~\ref{fig_genre1} we analyze the accuracy of kernel methods as a function of machine size. As before, it is clear that rkOPLS and rkCCA are the best performing methods both at the AR and song levels. Increasing the size of the machine results in better performance, although the improvement is not very noticeable in excess of 250 nodes. Altogether, these results allow us to conclude that sparse methods can be used to enhance the applicability of kMVA for large data sets. In this subsection, we have focused on the sparse-promotion technique of~\cite{ArenasGarcia07}, but similar advantages can be expected from other sparse approaches.

\section{Conclusions}\label{sec:conclusions}

We reviewed the field of supervised feature extraction from the unified framework of multivariate analysis. The use of these techniques in real world applications is becoming increasingly popular. Beyond standard PCA, there is a plethora of linear and kernel methods that are generally better suited for supervised applications since they seek for projections that maximize the alignment with the target variables. We analyzed the commonalities and basic differences of the most representative MVA approaches in the literature, as well as the relationships to existing kernel discriminative feature extraction and statistical dependence estimation approaches. We also studied recent methods to make kernel MVA more suitable to real life applications, both for large scale data sets and for problems with few labeled data. In such approaches, sparse and semi-supervised learning extensions have been successfully introduced for most of the models. Actually, seeking for the appropriate features that facilitate classification or regression cuts to the heart of manifold learning. We have illustrated MVA methods in challenging real problems that typically exhibit complex manifolds. A representative subset of the UCI repository has been used to illustrate the general applicability of the standard MVA implementations in moderate data sizes. We have completed the panorama with challenging real-life applications: the prediction of music genre from recorded signals, and the classification of land-cover classes and estimation of climate variables to monitor our Planet. This tutorial presented the framework of kernel multivariate methods and outlined relations and possible extensions. The adoption of the methods in many disciplines of science and engineering is nowadays a fact. New exciting advances in the theory and applications are yet to come.

\section*{Acknowledgments}
The authors would like to thank the reviewers and the special issue editors for useful comments that improved the presentation of the present manuscript significantly.

This work was partially supported by Banco Santander and Universidad Carlos III de Madrid's Excellence Chair programme, and by the Spanish Ministry of Economy and Competitiveness (MINECO) under projects TIN2012-38102-C03-01, TEC2011-22480 and PRI-PIBIN-2011-1266.

\bibliographystyle{IEEEbib}

\begin{thebibliography}{}

\end{thebibliography}


\begin{thebibliography}{10}
\bibitem{Wold66b}
H. Wold,
\newblock ``Estimation of principal components and related models by iterative least squares,''
\newblock in P. R. Krishnaiah (ed.) {\em Multivariate Analysis}, pp. 391--420, Academic Press, 1966.

\bibitem{Rosipal11}
R. Rosipal,
\newblock ``Nonlinear Partial Least Squares: An Overview,''
\newblock in {\em Chemoinformatics and Advanced Machine Learning Perspectives: Complex Computational Methods and Collaborative Techniques}, H. Lodhi and Y. Yamanishi (eds.), ACCM, IGI Global, pp. 169--189, 2011.

\bibitem{Woldetal89}
S. Wold, N. Kettaneh-Wold, and B. Skagerberg,
\newblock ``Nonlinear PLS Modeling,''
\newblock {\em Chemometrics and Intelligent Laboratory Systems}, vol. 7, pp. 53--65, 1989.

\bibitem{Qin92}
S. Qin and T. McAvoy,
\newblock ``Non-linear PLS modelling using neural networks,''
\newblock {\em Computers \& Chemical Engineering}, vol. 16, pp. 379--391, 1992.

\bibitem{Boser92}
B. E. Boser, I. Guyon, and V. N. Vapnik,
\newblock ``A training algorithm for optimal margin classifiers,''
\newblock in {\em Proc. COLT'92}, Pittsburgh, PA, 1992, pp. 144--152.

\bibitem{scholkopf1998}
B. Schölkopf, A. Smola and K.-R. Muller.
\newblock ``Nonlinear Component Analysis as a Kernel Eigenvalue Problem,''
\newblock {\em Neural Computation}, vol. 10, pp. 1299--1319, 1998.

\bibitem{ShaweTaylor04}
J. Shawe-Taylor and N. Cristianini,
\newblock {\em Kernel Methods for Pattern Analysis}, Cambridge University Press, 2004.

\bibitem{SPM11}
{\em IEEE Signal Process. Mag.,} special issue on Dimensionality Reduction via Subspace and Submanifold Learning, vol. 28, Mar. 2011.

\bibitem{Jollife86}
I. T. Jollife,
\newblock {\em Principal Component Analysis}, Springer-Verlag, 1986.

\bibitem{pearson1902} K. Pearson, ``On lines and planes of closest fit to systems of points in space,'' {\em Philosophical Mag.}, vol. 2 pp. 559--572, 1901.

\bibitem{Braun08}
M.~L.~Braun, J. M. Buhmann, and K.-R. Müller,
\newblock ``On Relevant Dimensions in Kernel Feature Spaces,''
\newblock {\em Journal of Machine Learning Research}, vol. 9, 1875--1908, 2008.

\bibitem{Abrahansem11}
T. J. Abrahamsen and L. K. Hansen,
\newblock ``A Cure for Variance Inflation in High Dimensional Kernel Principal Component Analysis,''
\newblock {\em Journal of Machine Learning Research}, vol. 12, 2027--2044, 2011.

\bibitem{wold1966a}
H. Wold,
\newblock ``Non-linear estimation by iterative least squares procedures,''
\newblock in F. David (ed.) {\em Research papers in Statistics}, pp. 411--444, New York, NY: Wiley, 1966.

\bibitem{wold1984} S. Wold, et al., ``Multivariate Data Analysis in Chemistry,'' in B. R. Kowalski (ed.) {\em Chemometrics, Mathematics and Statistics in Chemistry}, pp. 17--95. Reidel Publishing Company, Holland, 1984.

\bibitem{geladi1988} P. Geladi, ``Notes on the history and nature of partial least squares (PLS) modelling'', {\em Journal
of Chemometrics}, vol. 2, pp. 231--246, 1988.

\bibitem{kramer2006} N. Kramer and R. Rosipal, ``Overview and recent advances in partial least squares,'' in C. Saunders et al. (eds.) {\em Subspace,
Latent Structure and Feature Selection Techniques}, pp. 34--51, Springer-Verlag, 2006.

\bibitem{Hotelling36}
H. Hotelling,
\newblock ``Relations between two sets of variates,''
\newblock {\em Biometrika}, 28, 321--377, 1936.

\bibitem{Borga97}
M. Borga, T. Landelius, and H. Knutsson,
\newblock ``A unified approach to PCA, PLS, MLR and CCA,''
\newblock {\em Tech. Report LiTH-ISY-R-1992,} Link{\"o}ping, Sweden, 1997.

\bibitem{Barker03}
M. Barker and W. Rayens,
\newblock ``Partial least squares for discrimination,''
\newblock {\em Journal of Chemometrics}, vol. 17, pp. 166--173, 2003.

\bibitem{Zou06}
H. Zou, T. Hastie, and R. Tibshirani,
\newblock ``Sparse Principal Component Analysis,''
\newblock {\em Journal of Computational and Graphical Statistics}, vol. 15, pp. 265--286, 2006.

\bibitem{Hardoon11}
D. R. Hardoon and J. Shawe-Taylor,
\newblock ``Sparse Canonical Correlation Analysis,''
\newblock {\em Machine Learning}, vol. 83, pp. 331--353, 2011.

\bibitem{Heskes12}
M. van Gerven, Z. Chao, and T. Heskes
\newblock ``On the decoding of intracranial data using sparse orthonormalized partial least squares,''
\newblock {\em Journal of Neural Engineering}, vol. 9, 2012.

\bibitem{Bishop95}
C. M. Bishop,
\newblock {\em Neural networks for pattern recognition}, Oxford University Press, 1995.

\bibitem{lai2000} 
P. L. Lai and C. Fyfe,
\newblock ``Kernel and non-linear Canonical Correlation Analysis,''
\newblock {\em Intl.  Journal of Neural Systems}, vol. 10, pp. 365--377, 2000.

\bibitem{Rosipal01}
R. Rosipal and L. J. Trejo, 
\newblock ``Kernel partial least squares regression in reproducing Hilbert spaces,''
\newblock {\em Journal of Machine Learning Research}, 2, 97--123, 2001.

\bibitem{ArenasGarcia07}
J. Arenas-García, K. B. Petersen, and L. K. Hansen,
\newblock ``Sparse kernel orthonormalized PLS for feature extraction in large data sets,''
\newblock in {\em NIPS}, 19, MIT Press, 2007.

\bibitem{Biessmann10}
F. Biessmann, et.al.,
\newblock ``Temporal kernel CCA and its application in multimodal neuronal data analysis,''
\newblock {\em Machine Learning,} vol. 79, pp. 5--27, 2010.

\bibitem{Blaschko11}
M. Blaschko, J. Shelton, A. Bartels, C. Lampert, and A. Gretton, 
\newblock ``Semi-supervised Kernel Canonical Correlation Analysis with Application to Human {fMRI}'', 
\newblock {\em Patt. Recogn. Lett.} vol. 32, pp.1572--1583, 2011.

\bibitem{Mika99}
S. Mika, G. Rätsch, J. Weston, B. Schölkopf, and K.-R. Müller,
\newblock ``Fisher discriminant analysis with kernels,''
\newblock in {\em Proc. IEEE Neural Networks for Signal Processing Workshop}, Madison, WI, Aug. 1999, pp. 41--48. 

\bibitem{Mika03}
S. Mika et al.,
\newblock ``Constructing Descriptive and Discriminative Nonlinear Features: Rayleigh Coefficients in Kernel Feature Spaces,''
\newblock {\em IEEE Trans. Patt. Anal. Mach. Intell.}, vol. 25, pp. 623--628, 2003.

\bibitem{Baudat00}
G. Baudat and F. Anouar, 
\newblock ``Generalized Discriminant Analysis Using a Kernel Approach,''
\newblock {\em Neural Computation,} vol. 12, pp. 2385--2404, 2000.

\bibitem{Gretton05COLT}
A. Gretton, O. Bousquet, A. Smola, and B. Schölkopf,
\newblock ``Measuring statistical dependence with Hilbert-Schmidt norms,''
\newblock in {\em Proc. 16th Intl. Conf. Algorithmic Learning Theory}, Springer, 2005, pp. 63--77.

\bibitem{Gretton05}
A. Gretton, R. Herbrich and A. Hyv\"arinen,
\newblock ``Kernel methods for measuring independence'',
\newblock {\em Journal of Machine Learning Research}, vol. 6, 2075--2129, 2005.

\bibitem{Principe10}
J. C. Principe,
\newblock {\em Information Theoretic Learning: Renyi's Entropy and Kernel Perspectives}, Springer, 2010.

\bibitem{hoegaerts2005} 
L. Hoegaerts, J. A. K. Suykens, J. Vanderwalle, and B. De Moor,
\newblock ``Subset based least squares subspace regression in RKHS,''
\newblock {\em Neurocomputing}, vol. 63, pp. 293--323, 2005.

\bibitem{Tipping01}
M. E. Tipping,
\newblock ``Sparse Kernel Principal Component Analysis,''
\newblock in {\em NIPS}, 13, MIT Press, 2001.

\bibitem{Bennett03} 
M. Momma and K. Bennett,
\newblock ``Sparse kernel partial least squares regression,''
\newblock in {\em Proc. Conf. on Learning Theory}, 2003.

\bibitem{Dhanjal09}
C. Dhanjal, S. R. Gunn, and J. Shawe-Taylor,
\newblock ``Efficient sparse kernel feature extraction based on partial least squares,''
\newblock {\em IEEE Trans. Patt. Anal. and Mach. Intell.}, vol. 31, pp. 1347--1360, 2009.

\bibitem{Izquierdo12}
E. Izquierdo-Verdiguier, J. Arenas-Garc\'ia, S. Mu\~noz-Romero, L. G\'omez-Chova and G. Camps-Valls,
\newblock ``Semisupervised Kernel Orthonormalized Partial Least Squares,''
\newblock in {\em Proc. IEEE Mach. Learn. Sign. Proc. Workshop}, Santander, Spain, 2012.

\bibitem{ArenasGarcia08}
J. Arenas-García and G. Camps-Valls,
\newblock ``Efficient Kernel OPLS for remote sensing applications,''
\newblock {\em IEEE Trans. Geosc. Rem. Sens.}, 44, 2872--2881, 2008.

\bibitem{Meng07}
A. Meng, P. Ahrendt, J. Larsen, and L. K. Hansen,
\newblock ``Temporal Feature Integration for Music Genre Classification,''
\newblock {\em IEEE Trans. Audio, Speech and Lang. Process.}, vol. 15, pp. 1654--1664, 2007.

\bibitem{ArenasGarcia09}
J. Arenas-García and K. B. Petersen,
\newblock ``Kernel multivariate analsis in remote sensing feature extraction,''
\newblock in G. Camps-Valls and L. Bruzzone (eds.) {\em Kernel methods for Remote Sensing Data Analysis}, Wiley, 2009.

\bibitem{CampsValls12}
G. Camps-Valls, J. Mu\~noz-Marí, L. Gómez-Chova, L. Guanter, and X. Calbet,
\newblock ``Nonlinear Statistical Retrieval of Atmospheric Profiles from MetOp-IASI and MTG-IRS Infrared Sounding Data,''
\newblock {\em IEEE Trans. Geoscience and Remote Sensing}, 2012.

\end{thebibliography}

\end{document}